# Knowledge revision in systems based on an informed tree search strategy: application to cartographic generalisation

Patrick Taillandier
COGIT IGN
2/4 avenue Pasteur
94165 Saint-Mandé Cedex - France
Patrick.taillandier@gmail.com

Cécile Duchêne
COGIT IGN
2/4 avenue Pasteur
94165 Saint-Mandé Cedex - France
cecile.duchene@ign.fr

Alexis Drogoul
IRD, UR 079
FI/MSI, ngo 42 Ta Quang Buu, Hai
Ba Trung, Ha Noi, Viet Nam
alexis.drogoul@gmail.com

## ABSTRACT
Many real world problems can be expressed as optimisation problems. Solving this kind of problems means to find, among all possible solutions, the one that maximises an evaluation function. One approach to solve this kind of problem is to use an informed search strategy. The principle of this kind of strategy is to use problem-specific knowledge beyond the definition of the problem itself to find solutions more efficiently than with an uninformed strategy. This kind of strategy demands to define problem-specific knowledge (heuristics). The efficiency and the effectiveness of systems based on it directly depend on the used knowledge quality. Unfortunately, acquiring and maintaining such knowledge can be fastidious. The objective of the work presented in this paper is to propose an automatic knowledge revision approach for systems based on an informed tree search strategy. Our approach consists in analysing the system execution logs and revising knowledge based on these logs by modelling the revision problem as a knowledge space exploration problem. We present an experiment we carried out in an application domain where informed search strategies are often used: cartographic generalisation.

## Categories and Subject Descriptors
I.2.8 [**Artificial Intelligence**]: Problem Solving, Control methods and Search – *Graph and tree search strategies, Heuristic methods*

## 1. INTRODUCTION
Problem-solving is one of the central topics of artificial intelligence. Among solving approaches, some are based on an informed search strategy. The principle of this kind of strategy is to use problem-specific knowledge (heuristics) beyond the definition of the problem itself to find solutions more efficiently than with an uninformed strategy.

The efficiency of systems based on this kind of strategy directly depends on the used knowledge quality. Unfortunately, it is usually very difficult to acquire expert knowledge. Eward Feigenbaum formulated this problem in 1977 as the knowledge acquisition bottleneck problem. Indeed, the expert knowledge is rarely formalised and its translation into a formalism usable by computers is very complex.

The work presented in this paper deals with the problem of knowledge revision in systems based on a specific informed search strategy. We propose an approach of automatic knowledge revision for such systems.

In part 2, we introduce the general context in which our work takes place and the difficulties that we must face. Part 3 is devoted to the presentation of our approach. Part 4 describes an application of our approach to cartographic generalisation. In this context, we present a real case study that we carried out as well as its results. Part 5 concludes and presents the perspectives for this work.

## 2. CONTEXT
### 2.1 Context and formalisation of the revision problem

#### 2.1.1 Description of the considered optimisation problems
Many real world problems can be expressed as optimisation problems. The goal in this kind of problems is to find, among all possible solutions, the one that maximises an evaluation function.

In this paper, we are interested in a family of optimisation problems, which consist in finding, by action application, the state of an entity that maximises an evaluation function.

Let $P$ be an optimisation problem class that is characterised by:

- An entity class $E_P$
- $\{action\}_P$: a set of actions that can be applied on an entity belonging to $E_P$. The result of the application of an action is supposed non-predicable.
- $Q_P$: a function that defines the state quality of an entity belonging to $E_P$

An optimisation problem p of class P is defined by an entity $e_p$ of class $E_P$, which is characterised by its current state. Solving p consists in finding the state $s$ of $e_p$ that optimises $Q_P$, by applying actions from $\{action\}_P$ to the initial state of $e_p$.

Let's consider the following example: Let *Probot* be a class of problems where a robot, considering its initial position in a maze, seeks to find the.

- $E_{Probot}$ : a kind of robot. A robot of the kind $E_P$ is characterised by its initial position in the maze.

- *{action}$_{Probot}$*: *{move forward, turn left, turn right}*
- *$Q_{Probot}$*: distance separating the robot from the exit of a maze

A problem of the class *Probot* is: Let $e_{probot}$ be a robot of the kind $E_{Probot}$ with an initial position in the maze. Its goal is to find the exit or at least to reach the closest possible position to the exit.

There are many ways to solve problems of this kind. In this paper, we are interested in systems that solve it by exploring a state tree by means of an informed strategy. Such systems are often used for real world problems thanks to their efficiency. In section 2.1.2, we present the generic system for which our revision approach is dedicated. Our revision approach could be used for other kinds of systems with some adaptations.

### 2.1.2 Description of the considered systems
The generic system is based on informed depth-first exploration of state trees. The passage from a state to another corresponds to the application of an action. Figure 1 presents the action cycle.

It begins with the characterisation of the current state of the entity and its evaluation using the function $Q_P$. Then, the system tests if the current state is good enough or if it is necessary to continue the exploration of others states. If the system decides to continue the exploration, it tests if the current state is valid or not. If not, the entity backtracks to its previous state, otherwise, the system constructs a list of actions to apply. If the actions list is empty the entity backtracks to its previous state, otherwise the system chooses the best action, and applies it. Then it goes back to the first step. The action cycle ends when the stopping criterion is checked or when all actions have been applied for all valid states.

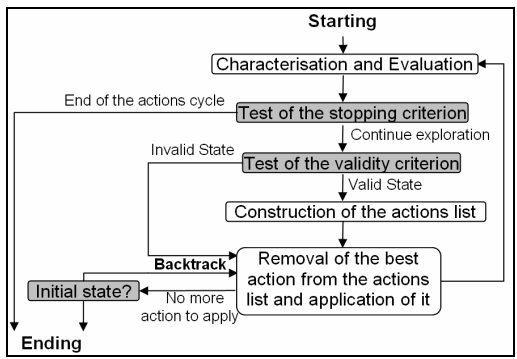

**Figure 1. Action cycle**

In this paper, we are interested in the pieces of procedural knowledge used to construct the actions list. One knowledge base $K_P$ is defined by optimisation problems class *P*. In many real world applications, knowledge is expressed by production rules. The interest of this kind of knowledge representation is to be easily interpretable by domain experts and thus to facilitate the knowledge validation and update. Therefore, we impose in our system the knowledge to be expressed by production rules.

A knowledge base $K_P$ contains, for each action, a production rule base that defines, for each state and according to a measure set, if the action has to be applied and if so, with which weight. The higher the weight, the higher priority the action has (and thus will be applied first). The weight is an integer between 0 and WEIGHT_MAX (0: the action is not proposed for the state, WEIGHT_MAX: the action is applied first). A measure set is defined per action. Several actions can depend of the same measure set. The advantage of having one rule base per action and not a unique rule base for all actions is to facilitate the definition of the rules by domain experts, to have more readable knowledge base and to improve the modularity of the system: it makes it easier to update the knowledge base when removing or adding new actions.

### 2.1.3 Formalisation of the knowledge revision
We define a function *perf(S,p)* that evaluates the performance of a system *S*, for the resolution of a problem *p*. The function is linked to the effectiveness and the efficiency of *S* for the resolution of *p* and depends on the domain and on the users needs.

We define, in the same way, a function *Perf(S,P)* that evaluates the performance of a system *S*, for the resolution of all problems of class *P*. The computation of the function demands to compute the function *perf(S,p)* for each *p* belonging to *P*. The knowledge revision problem then consists in finding for an optimisation problem class *P*, and with the help of the initial knowledge base, among all possible knowledge bases for *P*, the one that optimises *Perf(S,P)*.

In practice, most of the time, it is impossible to compute *Perf(S,P)*. Indeed, it is rarely possible to compute *perf(S,p)* for each *p* belonging to *P*. Thus, we will just estimate *Perf(S,P)* on a sample of problems of class *P*. The more representative of all problems of *P* the problems composing the sample, the better the estimation will be.

The choice of the sample has a major importance for the possible revision quality.

## 2.2 Related works
As we mention in part 2.1.2, we are interested in knowledge represented by productions rules. If many learning algorithms propose to induce production rules from examples labelled by experts, very few among them allow taking into account initial rules. However, a few works already dealt with this problem.

Among them, a few are interested in the inductive knowledge-base refinement. The objective of these works is to improve the expert system knowledge base. Most of them make the assumption that the knowledge base is almost valid and that only small improvements are needed [6]. Thus, some approaches propose to improve rule bases only by refining or deleting existing rules without giving the possibility to add new rules [9]. Others do not aim at refining rule bases directly, but aim at supporting the user during the refining process [1]. Many of these works are based on logical operators and thus rarely deal with noisy data [13]. Another drawback of many of these works is the increase of the number of rules [17] that can lead to readability problems. One common point of all these works is that they concern the revision of a unique rule base and do not allow revising several dependant rule bases simultaneously. It is thus not possible to directly apply these approaches to our revision problem.

## 3. PROPOSED APPROACH
## 3.1 General approach
Our objective is to automatically revise the knowledge base of a system based on an informed tree search strategy. The system already has a defined initial knowledge base that has to be taken into account by the revision process.

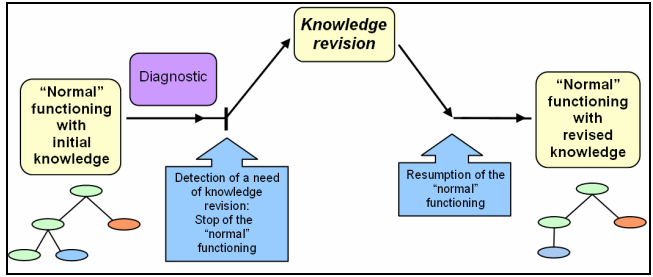

**Figure 2. General approach**

Our revision approach is based on the analysis of the execution logs. We do not seek on-line knowledge revision. Our approach requires to stop the "normal" system functioning in order to activate the process with a minimal pruning and thus get the most complete and accurate information on the successes/failures met by each piece of knowledge (figure 2).

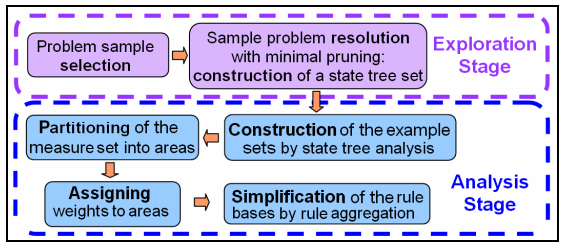

**Figure 3. Revision approach**

Our approach is composed of two stages (figure 3):

- *Exploration stage*: consists in logging the process while the system solves a great number of problems with minimal pruning. This stage is composed of two sub-stages:
  - *Selection* of the problem sample
  - *Resolution* of the selected problems and logging

- *Analysis stage*: consists in analysing the logs obtained during the previous stage and in using it to revise the knowledge. This stage is composed of four sub-stages:
  - *Construction* of example sets by analysis of the problems solved during the exploration stage
  - *Partitioning* of the measure sets values into areas using the example sets (and taking into account the initial knowledge base)
  - *Exploration* of the space of possible weights to affect to each area, in order to find the best
  - *Simplification* of the rule bases by rule aggregation

## 3.2 Exploration stage

During this stage, the system solves a sample of problems with a minimal pruning. The pruning is defined so that the system, for each problem, constructs all possible states according to the actions set and to the initial state of the problem while ensuring the action cycle converge. In this way, it is be possible to simulate each possible knowledge base by rearranging the states, without having to run again the system with this knowledge set.

This minimal pruning is chosen so the results obtained with it are independent of the initial pieces of knowledge: whichever the initial pieces of knowledge are, the state tree obtained with the minimal pruning is composed of the same states (which could be set in a different order).

Concerning the choice of the problem sample, we use a sampling method that is not developped in this paper.

## 3.3 Analysis stage

### 3.3.1 Proposed approach

During the analysis stage, the system revises the knowledge base thanks to the problems solved during the exploration stage. We propose to formulate the revision problem itself as a search problem. We will then search the knowledge base that optimises the function *Perf(S,$P_n$)* defined in part 2.1.3, with $P_n$, the sample of n problems solved during the exploration stage.

Let us remind that we deal with the revision of the action application knowledge. Each action has a rule base, which defines the weight of the action for each value of its measure set (see part 2.1.2). The difficulty comes from the distributed nature of this knowledge. Actually, if the application rule bases of each action are not dependent on each other in their expression (each action has a rule base which only depends on its own measure set), the results (the weight) can only be analysed if compared to the weight of the other actions. In fact, Given a system which has two actions $A_1$ and $A_2$ to solve a class of problems, knowing that, for a given entity state, the weight of $A_1$ is equal to 4 has no meaning if we do not know that for the same entity state, $A_2$ has a weight of 3. Therefore, it is not possible to revise the knowledge of each action independently: we have to take into account all actions at the same time.

In order to reduce the search space, we propose to decompose the measure set space into areas by partitioning it while taking into account the initial rules and information obtained from solved problems. For example, consider a system that can propose only the action *A* that depends on a measure set composed of just one real measure *M*. An example of partitioning can be to decompose the domain of *M* (and thus the measure set space of *A*) into two areas: (M < 0) and (M ≥ 0).

The revision problem then consists in assigning, for each action, the best possible weight to each area of its measure set. We call solution, a complete assignment, for each action, of weights for each area of its measure set. After finding the best solution, it is possible to simplify the resulting rule bases by rule aggregation.

The partitioning of the measure sets is based on the results obtained from the solved problems during the exploration stage. In fact, we build example sets from the solved problems and we use them for the measure sets decomposition. The next part introduces the construction of the example sets.

### 3.3.2 Construction of the example sets

The construction of the example sets is achieved by the analysis of the state trees obtained during the exploration stage. An example set is build per action. For each action, an example is composed of a state and is labelled with a "decision". A state is described by the measure set linked to the action. A "decision" is either a "success" or a "failure". We define the notion of best path: a best path is a sequence of at least two states, which has the root of a tree (or of a sub-tree) for initial state and the best state of this tree (or sub-tree) for final state. The construction of our example set for a state tree consists, in a first step, in extracting the best paths from the tree. The next step consists in analysing each state of each best path. If, from one of these states, one of the actions proposed leads to another state of the same best path, the action is noted as having a success. Otherwise, it is noted as having a failure. Figure 4 gives a simplified example of the example sets built from the resolution of a problem $p$ with two actions $A_1$ and $A_2$.

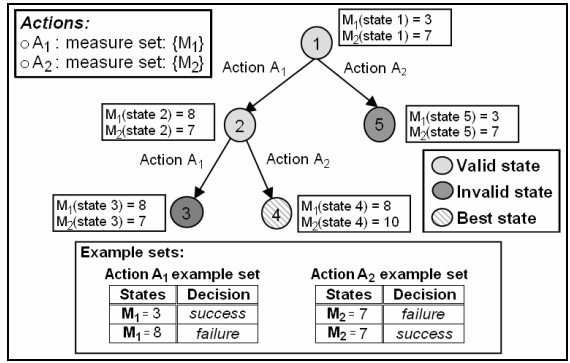

**Figure 4. Example of a built example sets**

### 3.3.3 Partitioning of the measure set space

Our general revision approach requires, as a first step, to partition the measure set into disjoint areas. The areas are defined by production rules.

One constraint of this partitioning is to take into account the initial action application rules. We impose for that, that each rule defining a partition of the measure set space must be either one the initial rules or either a specialisation of one of the initial rules. The interest of this constraint is to keep the possibility to obtain rules similar to the initial rules after the revision process. Several approaches based on the utilisation of the example sets, can be used to solve this problem.

The partitioning approach that we propose consists in discretising each measure, and in recomposing rules while taking into account the initial rules (the measures used and their cut). For example, consider an action $A$ which is linked to two measures $M_1$ and $M_2$. Let the initial rules be:

if ($M_1 < 5$) then weight = 2
if ($M_1 \geq 5$) and ($M_2 < 3$) then weight = 1
if ($M_1 \geq 5$) and ($M_2 \geq 3$) then weight = 0

If the discretisation algorithm decomposes $M_1$ in two areas $]-\infty;0]$ and $]0;\infty[$ and $M_2$ into one area $]-\infty, \infty[$, the resulting rules would be :

if ($M_1 \leq 0$) then weight = 2
if ($M_1 < 0$) and ($M_1 < 5$) then weight = 2
if ($M_1 \geq 5$) and ($M_2 < 3$) then weight = 1
if ($M_1 \geq 5$) and ($M_2 \geq 3$) then weight = 0

Others approaches can be used for the partitioning. For example, it is possible to use a supervised learning algorithm to decompose the measure set to specialise them a posteriori by comparing them to the learnt rules.

Once the partitioning carried out, the revision process consists in assigning, for each area of each measure set, the best possible weight.

### 3.3.4 Exploring stage

We defined our revision problem as an optimisation problem in which we search, for a given problem class $P$ and a given system $S$, the solution $sol$ among the possible solutions set $Sol$, that maximises the quality function $Perf(S_{sol}, P_n)$. According to the fact that for each area, we have to assign a weight value between 0 and WEIGHT _MAX, the size of the solutions space (size of $Sol$) is equals to $(1+ \text{WEIGHT \_MAX})^{\text{number of areas}}$.

To help this search, we dispose of an initial solution (the initial knowledge base) that will often be good. There are numerous methods to solve a problem of this kind. Due to the size of the solution space, it is impossible to use a complete approach. Thus, we use an incomplete approach. Indeed, in order to solve this problem, we propose to use a reactive local search algorithm [3]. Others algorithms such as hill climbing, tabu search [10], simulated annealing [11], can also be used to solve this problem. The principle of this kind of algorithm is to start with an initial solution and try to improve it by exploring its neighbourhood. They are, most of the time, very effective for this kind of exploration problem.

Local search approaches require defining a notion of neighbourhood for a solution. For our problem, it means the set of solutions for which only one of the areas will have its weight value changed with a neighbour value. For a given weight $W$, the neighbour weights are $W + 1$ and $W - 1$.

### 3.3.5 Rule base simplification

The exploring stage allows the system to assign a good weight to each area. The last step of the analysis stage consists in simplifying the obtained (revised) rules bases by aggregating the rules.

For example, if the resulting weight assignment for an action is the following:

if ($M_1 < 5$) then weight = 1
if ($M_1 \geq 5$) and ($M_2 < 3$) then weight = 3
if ($M_1 \geq 5$) and ($M_2 \geq 3$) then weight = 3,

the last two rules are aggregated and the final rule base is:

if ($M_1 < 5$) then weight = 1
if ($M_1 \geq 5$) then weight = 3

## 4. APPLICATION TO CARTOGRAPHIC GENERALISATION

### 4.1 Automatic cartographic generalisation

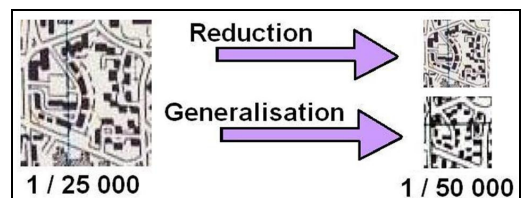

**Figure 5. Cartographic Generalisation**

Cartographic generalisation a process that aims at decreasing the level of details of geographic data in order to produce a map at a given scale. Figure 5 gives an example of cartographic generalisation. As illustrated in the figure, cartographic generalisation is not a simple size reduction. The application of numerous operations such as local scaling, displacements or elimination of objects are needed in order to ensure the readability of the map while keeping the essential information of the initial map.

The automation of this process is a complex problem, which has been the core of numerous research works in the recent years. Some of these works try to solve it by a local, step-by-step and knowledge-based approach [4, 18]. The difficulty then consists in choosing the best sequence of generalisation operations to apply on the various geographic objects. An approach to solve this problem is to use an informed search. Nowadays, the procedural knowledge used to guide the search is entered "by hand" by generalisation experts. Its tuning is often long and fastidious because it demands to face the problem of knowledge collecting and formalizing [18]. Several works have already used machine learning to learn the relevant procedural knowledge [12, 18] but few among them propose to automatically revise existing knowledge. One of them is [5] that proposes to use previously generalised objects to build a case base. Concerning the rule base revision, the only work existing is [15]. It proposes to use experience to learn new rules that are added in the system. This work does not propose to revise the existing rules, but just to add new ones.

The automation of cartographic generalisation is a particularly interesting industrial application context. In fact, first, it is a problem which is far from being solved. Moreover, it directly concerns many mapping agencies that wish to improve their map production lines. Finally, it touches the problem of on-demand mapping that takes a more and more important place with the multiplication of the possibilities to create one's own map on the web.

### 4.2 The generalisation system

The generalisation system that we use for our experiment is based on the AGENT model [2, 14] and follows the specification that we defined in part 2.2.2. It generalises a geographic object or a group of geographic objects by the mean of an informed tree search strategy.

Each state represents the geometric state of the considered geographic objects and is evaluated by a *satisfaction* function, which translates the respect of cartographic constraints by the geographic objects. A cartographic constraint can be for a building to be big enough to be readable. The satisfaction of a state is ranged between 1 and 10 (10 represents a perfect state and a score lower than 5, a non acceptable state). The actions cycle used is the one presented figure 1. The stopping criterion is the following: the action cycle ends when the system reaches a perfect state (or when all possible states have been visited). The validity criterion depends of the cartographic constraints satisfaction improvement. The weight of the actions is ranged between 0 and 5.

### 4.3 Application of our revision approach

We applied our revision approach to revise the action application knowledge of our generalisation system.

Concerning the partitioning of the measure sets, we used the algorithm proposed by [8] to discretise the measures.

We chose the reactive local search algorithm [3] for our search problem.

The function $Perf(S_K, P_n)$ defined is the following:

$$Perf(S_K, P_n) = \frac{1}{4}\left(3 \times \frac{Mean\ Satisfaction}{10} + \frac{1}{\sqrt{Mean\ Nb\ of\ States}}\right)$$

The mean satisfaction represents the effectiveness of the system. The higher the mean satisfaction, the more effective system is. The mean number of states represents the efficiency of the system. The higher the mean number of states, the less efficient the system is. The function $Perf(S_K, P_n)$ is ranged between 0 (very bad results) and 1 (perfect results). We can explain this formula by the fact that the satisfaction is ranged between 1 and 10 and the number of state between 1 and $\infty$. In order to favour the effectiveness of the system rather than its efficiency, we introduce a factor 3 in favour of the satisfaction.

### 4.4 Case study

The real case study that we carried out concerned the generalisation of geographic object of the kind "building group". The building group generalisation is an interesting case study because it is not yet well mastered and because it is very time consuming.

We defined five actions for the building group generalisation as well as two knowledge bases: the first one is defined by a generalisation expert ($K_{expert}$), the second one corresponds to the case where no actions are proposed for any state ($K_{noAction}$). The "expert" knowledge base revision corresponds to the classical scenario of knowledge revision where we have a good initial rule base that we want to refine. The revision of the "no action" knowledge base corresponds to the scenario where the initial knowledge base is the worst possible and where we want to acquire good knowledge to replace it.

50 building groups were automatically selected among more than 300 available for the revision process (the learning sample). We tested the initial and the revised knowledge on a different area than the one used for the revision (the test sample). The area used for the test was composed of 155 building groups.

### 4.5 Results

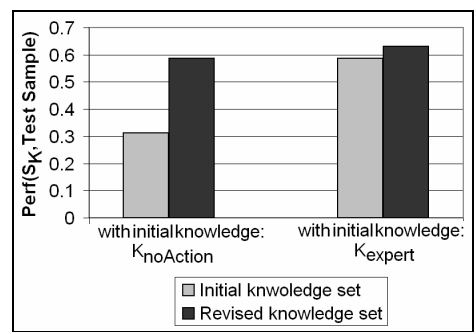

**Figure 6. Revision results**

The results of this experiment (figure 6) show that our revision approach improved the system knowledge. In fact, with both $K_{expert}$ and $K_{noAction}$ as initial knowledge, the results obtained with revised knowledge are better than the ones obtained with the initial knowledge base. These results validate our general approach.

An interesting point concerns the way the initial knowledge base is taken into account. The revised knowledge obtained from the revision of $K_{expert}$ obtained better results than the one obtained from the revision of $K_{noAction}$. An explanation for that is that the expert integrated, in its knowledge base, information that was not present in the learning sample and that was kept by the revision process. This results show the interest of taking into consideration the initial knowledge base for the revision process.

Figure 7 gives examples of building groups generalised with the different knowledge bases. These examples show that the generalisation obtained with both revised knowledge bases are better than the ones obtained with both initial knowledge bases.

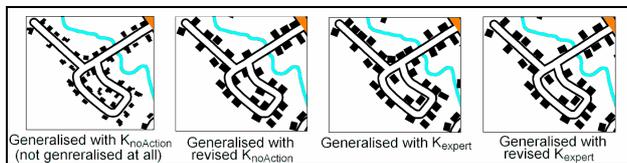

**Figure 7. Building group generalisation examples**

## 5. CONCLUSION

In this paper, we proposed a knowledge revision approach based on the exploration of the knowledge space. We showed the effectiveness and the efficiency of our approach on a real case study.

If we revised the action application knowledge, we did not try to revise others pieces of knowledge like the validity criterion or the actions cycle ending criterion. Some adaptation of our approach could be proposed to revise as well this kind of knowledge. In the same way, adaptations could be proposed to revise knowledge expressed in others formalisms than production rules.

A point that deserves more study is the problem sample choice. In fact, depending of the choice, the revision results that can be obtained can be very different in quality. Another point that deserves more study is the knowledge space partitioning. A bad partitioning does not allow to improve the initial knowledge.